\documentclass[journal]{IEEEtran}

\usepackage{cite}
\usepackage{amsmath,amssymb,amsfonts}
\usepackage{algorithmic}
\usepackage{algorithm}
\usepackage{graphicx}
\usepackage{textcomp}
\usepackage{xcolor}
\usepackage{booktabs}
\usepackage{multirow}
\usepackage{float}

\def\BibTeX{{\rm B\kern-.05em{\sc i\kern-.025em b}\kern-.08em
    T\kern-.1667em\lower.7ex\hbox{E}\kern-.125emX}}
\begin{document}

\title{Adaptive Sequential Test Planning for Multi-Mechanism Reliability Qualification via Bayesian Monte Carlo Tree Search}

\author{Youssef A. Elhagrasy, Ian Hill,~\IEEEmembership{Member,~IEEE,} and André Ivanov,~\IEEEmembership{Fellow,~IEEE}%
\thanks{Manuscript received Month Day, Year; revised Month Day, Year.}%
\thanks{Corresponding author: Youssef A. Elhagrasy (e-mail: yelhagra@student.ubc.ca).}
\thanks{Y. A. Elhagrasy, I. Hill, and A. Ivanov are with the Department of Electrical and Computer Engineering, University of British Columbia, Vancouver, BC V6T 1Z4, Canada.}
\thanks{© 2026 IEEE. Personal use of this material is permitted. Permission from IEEE must be obtained for all other uses, in any current or future media, including reprinting/republishing this material for advertising or promotional purposes, creating new collective works, for resale or redistribution to servers or lists, or reuse of any copyrighted component of this work in other works.}%
}

\markboth{IEEE Transactions on Reliability,~Vol.~XX, No.~X, Month~Year}%
{Elhagrasy \MakeLowercase{\textit{et al.}}: Adaptive Sequential Test Planning via Bayesian MCTS}

\maketitle

\begin{abstract}
Reliability qualification of advanced semiconductor devices requires sequential stress decisions that balance characterization objectives against multiple competing failure mechanisms. Current practice relies on static test plans derived from population-level acceleration models, which cannot adapt to per-unit variability or real-time degradation observations. This paper presents a closed-loop adaptive test planning framework that formulates reliability qualification as a partially observable sequential decision problem and solves it using Monte Carlo tree search for seed-action simulators (MCTS-SA) coupled with extended Kalman filter (EKF) belief-state estimation. The framework models stochastic, per-device variability in bias temperature instability (BTI), electromigration (EM), and time-dependent dielectric breakdown (TDDB), and treats stress selection as a constrained sequential optimization, i.e., to maximize the probability of successful degradation characterization while respecting catastrophic failure constraints. Under the experimental assumptions used here (discrete stress actions, proxy damage observability, and cumulative degradation without recovery), we believe this to be a novel application of tree-search-based adaptive test planning to multi-mechanism reliability qualification. Across 5{,}000 planning iterations, the characterization yield (\textbf{CY}) improves from 20\% in the first 500 iterations to over 54\% in the final 500, with 39\% cumulative success, while the best successful test sequence terminates with EM and TDDB damage fractions $D_{\mathrm{EM}}=0.564$ and $D_{\mathrm{TDDB}}=0.537$, well within safety margins. These results demonstrate that sequential Bayesian planning can synthesize damage-aware test policies that significantly outperform non-adaptive strategies for reliability qualification under competing failure modes.

\end{abstract}


\section*{Abbreviations \& Acronyms}
\begin{tabular}{@{}ll}
ALT & Accelerated Life Test \\
AST & Adaptive Stress Testing \\
BTI & Bias Temperature Instability \\
CY & Characterization Yield \\
DUT & Device Under Test \\
EKF & Extended Kalman Filter \\
EM & Electromigration \\
FinFET & Fin Field-Effect Transistor \\
JEDEC & Joint Electron Device Engineering Council \\
MCTS & Monte Carlo Tree Search \\
MCTS-SA & MCTS for Seed-Action Simulators \\
MTTF & Mean Time To Failure \\
POMDP & Partially Observable Markov Decision Process \\
RL & Reinforcement Learning \\
TDDB & Time-Dependent Dielectric Breakdown \\
UCB & Upper Confidence Bound \\
UKF & Unscented Kalman Filter \\
\end{tabular}

\section{Introduction}
Reliability qualification of integrated circuits is fundamentally a sequential decision problem. At each stage of an accelerated test, an expert decision must be made to choose stress conditions that yield maximal information about device lifetime while avoiding premature failure through unintended mechanisms. At advanced semiconductor technology nodes, three dominant wear-out mechanisms, bias temperature instability (BTI), electromigration (EM), and time-dependent dielectric breakdown (TDDB), exhibit comparable timescales under accelerated conditions~\cite{pae2015finfet,rahman2018reliability,weir2019finfet}, creating a multi-objective optimization landscape where aggressive stress accelerates the target characterization but increases the risk of catastrophic failure through competing mechanisms.

We use BTI threshold-voltage shift $\Delta V_t$ as the observable proxy because it is the standard measurable signature of BTI degradation in stress-measure-stress characterization, and because BTI exhibits stress-induced drift and partial recovery that make the threshold shift more informative than an abstract latent-damage scalar for closed-loop planning~\cite{grasser2011bti,mahapatra2013bti}. In this paper, $\Delta V_t$ is the measured progress variable, while the underlying BTI parameters remain latent and are inferred {\it online}, i.e., the belief state over the parameters is updated after each measurement and used to inform the subsequent stress decision.

Current qualification practice, codified in JEDEC standards~\cite{jedec2016}, 
relies on static test plans—fixed stress recipes derived from 
population-average acceleration models and applied uniformly to all devices. 
While reproducible and well-understood, such plans exhibit three key limitations 
for advanced-node devices:
\begin{enumerate}
    \item They cannot incorporate real-time measurement data to adapt 
    subsequent stress decisions.
    
    \item They treat all devices identically, ignoring variability in 
    latent degradation parameters.
    
    \item They do not explicitly account for competing failure mechanisms, 
    instead relying on conservative stress margins that may either risk 
    unintended failure or prolong test time.
\end{enumerate}

The reliability engineering community has long recognized the value of 
sequential experimental design~\cite{nelson2004accelerated,meeker1998statistical,chaloner1995boed}, yet closed-loop, feedback-driven test planning for multi-mechanism qualification remains an open problem.

Here we address this gap by formulating reliability qualification as a constrained partially observable Markov decision process (POMDP)~\cite{kaelbling1998pomdp} and solving it with Monte Carlo tree search for seed-action simulators (MCTS-SA)~\cite{lee2018ast,corso2021survey}, a planning approach that is well matched to expensive stochastic simulation. We define a successful outcome as a \emph{characterization event}, where sufficient target degradation (BTI threshold-voltage shift $\Delta V_t \geq \Delta V_{\text{fail}}$) is induced without triggering catastrophic EM or TDDB failure, and refer to the rate of such events as the characterization yield (\textbf{CY}). An extended Kalman filter (EKF)~\cite{thrun2005probabilistic} maintains Bayesian belief states over the latent BTI parameters, providing the uncertainty quantification needed for informed test planning.

The key methodological insight is that reliability qualification under competing failure modes constitutes a constrained exploration problem amenable to a tree search where the planner must discover test sequences that navigate toward informative degradation trajectories while remaining within the safe operating region bounded by catastrophic failure surfaces. MCTS is well-suited to this setting because it handles sparse, non-convex reward landscapes without requiring differentiable objectives~\cite{browne2012mcts}, and the seed-action formulation~\cite{lee2018ast,corso2021survey} enables deterministic replay of stochastic simulations, a critical property for consistent planning with computationally expensive physics-based simulators.

This work is, to our knowledge, the first to leverage tree-search-based sequential planning for multi-mechanism reliability qualification. Its primary contributions are:
\begin{enumerate}
    \item A formulation of reliability qualification as a constrained sequential decision problem with competing failure modes, Bayesian belief-state dynamics, and a multi-component objective balancing \textbf{CY}, uncertainty reduction, and catastrophe avoidance.
    \item A stochastic reliability simulation environment modeling concurrent BTI, EM, and TDDB degradation with per-device latent variability, implemented using an existing physics-based temporal simulator~\cite{hill2023gerabaldi} and calibrated to exhibit realistic mechanism competition consistent with published industrial data~\cite{pae2015finfet,rahman2018reliability,weir2019finfet}.
    \item Integration of MCTS-SA with online EKF inference for adaptive test planning, with experimental validation demonstrating progressive improvement in characterization success rate across 5{,}000 planning iterations and stable convergence of the search tree.
    \item Empirical analysis of the planning algorithm's convergence behavior, including tree diagnostics, ablation studies of the reward structure, and robustness assessment across multiple configurations.
\end{enumerate}

The remainder of this paper is organized as follows. Section~II reviews related work in sequential test planning, adaptive stress testing, and Bayesian experimental design. Section~III formalizes the qualification problem. Section~IV describes the stochastic simulation environment. Section~V details the reward structure. Section~VI presents the MCTS-SA planning framework. Section~VII reports experimental results, and Section~VIII discusses implications, limitations, and conclusions.

\section{Related Work}
The reliability engineering literature has a rich tradition of optimal test planning. Nelson~\cite{nelson2004accelerated} established foundational methods for accelerated life test (ALT) design, including optimal allocation of test units across stress levels to minimize variance in lifetime estimates. Meeker and Escobar~\cite{meeker1998statistical} extended these methods to censored data and multi-stress settings. Chaloner and Verdinelli~\cite{chaloner1995boed} provided a comprehensive framework for Bayesian experimental design where test conditions are chosen to maximize expected information gain.

However, these classical methods are typically \emph{open-loop} in that the test plan is designed once before testing begins and executed without modification. In contrast, our framework is \emph{closed-loop} such that each stress decision incorporates all prior measurement information, enabling the planner to adapt to the specific device under test. Wu {\it et al.}~\cite{wu-milor-optimal-tddb-tests} developed optimal accelerated test frameworks for TDDB lifetime parameter estimation, but their approach optimizes static stress allocation rather than sequential, feedback-driven decisions. The gap between static optimal design and adaptive sequential planning for the reliability of systems subject to multi-mechanism modes of failure motivates our work.

Adaptive stress testing (AST)~\cite{lee2018ast,koren2018ast} formulates failure discovery as a sequential decision process, originally developed for safety-critical autonomous systems where finding rare failure modes in simulation is essential before deployment. Corso {\it et al.}~\cite{corso2021survey} survey black-box safety validation algorithms, including AST variants with different search backends. Previous AST applications have focused on autonomous vehicles~\cite{koren2018ast,corso2020adaptive} and aircraft collision avoidance~\cite{lee2018ast}. The seed-action formulation~\cite{lee2018ast,corso2021survey} extends MCTS to stochastic simulators by treating random seeds as actions, enabling deterministic trajectory replay.

We adapt this framework to a fundamentally different domain. In our case, we focus on reliability qualification where (i)~degradation is irreversible, so stress decisions cannot be undone; (ii)~multiple competing mechanisms constrain the feasible stress space; and (iii)~the objective is to achieve successful characterization by safely navigating the risk of catastrophic failure, rather than seeking failure as the end goal.

Reinforcement learning (RL) formalizes sequential decision-making as an 
interaction loop in which an agent observes the system state, selects actions, and receives scalar rewards, with the objective of maximizing cumulative return~\cite{sutton2018reinforcement}. Two principal solution paradigms are commonly used. Policy gradient methods~\cite{schulman2017ppo} represent decision strategies using parameterized function approximators, typically neural networks, while tree search methods~\cite{browne2012mcts,kocsis2006uct,coulom2007mcts} construct decisions by explicitly exploring sequences of actions. Policy gradient methods are effective when environment interaction is inexpensive, but they struggle in settings with sparse and highly non-convex reward landscapes, as encountered in reliability testing. In contrast, Monte Carlo Tree Search (MCTS) has demonstrated strong performance in long-horizon planning under uncertainty, 
from game playing~\cite{silver2016alphago} to engineering design~\cite{williams2017mppi}. We therefore adopt MCTS for its ability to plan efficiently without requiring differentiable objectives.

Bayesian approaches enable uncertainty quantification over model parameters~\cite{thrun2005probabilistic}, which is critical for informed test planning~\cite{chaloner1995boed}. The extended Kalman filter (EKF) provides computationally efficient approximate Bayesian inference for nonlinear systems by linearizing the measurement model. While alternatives such as the unscented Kalman filter (UKF)~\cite{julier1997ukf} avoid linearization, we select the EKF for efficiency given the moderate nonlinearity of the degradation models in our scenario. Machine learning has also been applied to degradation modeling~\cite{hill2023gerabaldi}. In contrast, our approach performs closed-loop Bayesian inference during testing, maintaining a belief state over degradation parameters that is actively updated from measurements and used to guide sequential stress selection via planning.

\section{Problem Formulation}

We formulate reliability qualification as a constrained partially observable sequential decision problem. Each test sequence corresponds to a distinct device instance sampled from a population with latent variability~\cite{meeker1998statistical}.

\subsection{Sequential Decision Framework}
The qualification process evolves over discrete decision epochs $t \in \{0, \dots, T_{\max}\}$. At each epoch, the planner observes available information and selects the next stress condition. The latent device parameters $\boldsymbol{\ell} = (\boldsymbol{\theta}_{\text{BTI}}, \boldsymbol{\theta}_{\text{EM}}, \boldsymbol{\theta}_{\text{TDDB}})$ are sampled once at initialization and remain fixed, representing the true physical characteristics of the device under test (DUT).

\subsubsection{Information State}
The observable information state $s_t$ available to the planner includes:
\begin{itemize}
    \item Accumulated stress time $T_{\text{accum}}$ (hours)
    \item Measured threshold voltage shift $\Delta V_t^{\text{meas}}$
    \item Estimated degradation parameters $\hat{\boldsymbol{\theta}}_{\text{BTI}}$ with posterior uncertainty $\boldsymbol{\Sigma}$
    \item Normalized damage indices $D_{\text{EM}} \in [0,1]$ and $D_{\text{TDDB}} \in [0,1]$
\end{itemize}

Because the true BTI and competing-mechanism parameters are not directly observable, this constitutes a POMDP~\cite{kaelbling1998pomdp}. In a POMDP the planner cannot act on the true physical state; instead it acts on a \emph{belief state}, defined as the posterior distribution over the unobserved quantities given the entire history of past stress actions and measurements, $b_t(\cdot) = p(\,\cdot \mid a_{0:t-1}, y_{1:t})$. The belief state is a sufficient statistic for optimal decision-making under partial observability~\cite{kaelbling1998pomdp}: it summarizes all information in the measurement history needed to select the next action, so the planner need not store the raw history. Here the planner maintains a belief state over the BTI parameters using an EKF (Section~\ref{sec:ekf_estimation}), while EM and TDDB damage fractions serve as measurement-derived proxy indices rather than latent state variables (see Section~\ref{sec:observability} for discussion).

\subsubsection{Decision Variables}
At each epoch, the planner selects stress conditions $a_t = (V, J, T, \Delta t)$ specifying voltage, current density, temperature, and duration. Without loss of generality, to provide a concrete example of our proposed methodology, we quantize the decision space as
\begin{align}
V &\in \{0.9, 1.0, 1.1, 1.2\} \text{ V} \\
J &\in \{0.8, 1.0, 1.5, 2.0, 2.5, 3.0\} \text{ (normalized)} \\
T &\in \{325, 350, 375\} \text{ K} \\
\Delta t &\in \{50, 100, 150, 200\} \text{ hours}
\end{align}
where normalized current density is defined as $J = I/I_{\mathrm{ref}}$ with $I_{\mathrm{ref}}$ set by the nominal stress condition.

\subsection{Qualification Outcomes}
A test sequence terminates when one of the following outcomes occurs:
\begin{enumerate}
    \item {\it Characterization Success}: $\Delta V_t^{\text{meas}} \geq \Delta V_{\text{fail}}$ (threshold $0.09$ V, see rationale in next sub-section), indicating sufficient degradation for reliable parameter estimation and lifetime extrapolation.
    \item {\it Catastrophic Failure}: $D_{\text{EM}} \geq 1$ or $D_{\text{TDDB}} \geq 1$, indicating irreversible device damage that precludes further characterization.
    \item {\it Test Time Budget Exhausted}: $t = T_{\max}$ (300 epochs), indicating the allocated test budget is consumed without achieving characterization.
\end{enumerate}

The planner's objective is to maximize the probability of outcome~(1) while minimizing the probability of outcome~(2), subject to the finite test time budget constraint~(3). This multi-objective formulation is encoded via a shaped reward function (Section~\ref{sec:reward}).

\subsection{Observability Model}
\label{sec:observability} \label{sec:bti_threshold_justification}
The characterization threshold $\Delta V_{\text{fail}} = 0.09$ V represents the degradation level at which parameter estimation achieves sufficient precision for lifetime extrapolation. We use the absolute threshold shift, $\Delta V_t$, as the primary metric rather than a fractional change (e.g., $\Delta V_t / V_{t,0}$). The physical basis for BTI is the generation of interface traps ($N_{it}$) at the semiconductor-dielectric interface. To a first-order approximation, the absolute shift $\Delta V_t$ is directly proportional to the density of these newly created traps~\cite{alam2003nbti}. A fractional metric, by contrast, would normalize this physical quantity by the initial device characteristics ($V_{t,0}$), confounding the measurement of the degradation itself with baseline device parameters that are not relevant to the BTI degradation trajectory. Using the absolute $\Delta V_t$ thus provides a more direct and physically grounded measure of the degradation progress. This value corresponds to approximately 45\% of the asymptotic $\Delta V_{\max}$ (mean 0.20 V), placing trajectories in the nonlinear regime of the stretched-exponential curve where the model parameters $(\tau, \beta)$ become identifiable rather than weakly constrained~\cite{grasser2011bti,mahapatra2013bti}. Formally, the threshold is selected so that the posterior over $(\tau,\beta)$ is concentrated enough to support ranking and extrapolation, while still leaving margin before EM and TDDB acceleration consumes the remaining stress budget. The choice therefore reflects a three-way tradeoff among signal-to-noise ratio, parameter identifiability, and competing-mechanism headroom.
We note that $D_{\text{EM}}$ and $D_{\text{TDDB}}$ are treated as directly observable proxy indices in this work; the implications of this simplifying assumption are discussed in Section~VIII.
\subsection{Seed-Action Formulation for Deterministic Planning}
To enable deterministic planning with MCTS, we employ a seed-action simulator abstraction~\cite{lee2018ast,corso2021survey}. Each decision is represented as a pseudorandom seed $\xi \in \{0, \ldots, 2^{32}-1\}$ that deterministically maps to stress parameters:
\begin{equation}
a_t(\xi_t) = \text{seed\_to\_action}(\xi_t) = (V(\xi_t), J(\xi_t), T(\xi_t), \Delta t(\xi_t))
\end{equation}
Concretely, the seed initializes a deterministic pseudorandom generator that draws one index per decision variable, independently and uniformly, from the corresponding discretized set in (1)--(4):
\begin{equation}
\label{eq:seed_map}
\begin{aligned}
\xi_t &\rightarrow \mathrm{PRNG}(\xi_t) \rightarrow (V, J, T, \Delta t),\\
&\quad V = \mathcal{V}[i_V], \quad i_V \sim \mathcal{U}\{0, \dots, |\mathcal{V}|-1\},
\end{aligned}
\end{equation}
and analogously for $J$, $T$, and $\Delta t$, where $\mathcal{V}, \mathcal{J}, \mathcal{T}, \mathcal{D}$ denote the action sets in (1)--(4) and $\mathcal{U}$ a uniform integer draw. Because the generator is seeded by $\xi_t$ alone, the mapping is a pure function of the seed. This ensures that replaying the same seed sequence from a given state produces identical trajectories, enabling consistent planning statistics. This property is critical for reliability test planning where each simulation may be computationally expensive but must be exactly reproducible.

\section{Stochastic Multi-Mechanism Simulation Environment}

We implement the reliability environment using an existing physics-based temporal simulator~\cite{hill2023gerabaldi} which provides degradation modeling with configurable latent parameter distributions. This section summarizes the degradation models; the emphasis is on the stochastic structure that creates the planning challenge rather than on the device physics per se.

\subsection{Per-Device Latent Variability}
At test initialization, device-specific latent parameters are sampled from physically motivated distributions~\cite{mahapatra2013bti,meeker1998statistical} and remain fixed throughout the test. Table~\ref{tab:latent_distributions} summarizes the distributions for the three degradation mechanisms considered in this work, namely BTI, EM, and TDDB.
\begin{table}[!htbp]
\caption{Latent Parameter Distributions (Per Device)}
\label{tab:latent_distributions}
\centering
\setlength{\tabcolsep}{4pt}
\renewcommand{\arraystretch}{1.05}
\footnotesize
\begin{tabular}{l l p{0.30\columnwidth} p{0.23\columnwidth}}
\toprule
\textbf{Mechanism} & \textbf{Parameter} & \textbf{Distribution} & \textbf{Support/Bounds} \\
\midrule
BTI  & $\Delta V_{\max}$ & $\mathcal{N}(0.20,\,0.02^2)\ \mathrm{V}$ & $\Delta V_{\max}>0$ \\
BTI  & $\tau$            & $\mathcal{N}(28000,\,2800^2)\ \mathrm{hr}$ & $\tau>0$ \\
BTI  & $\beta$           & $\mathcal{N}(0.5,\,0.03^2)$ & $[0.15,\,0.95]$ \\
\midrule
EM   & $A_{\mathrm{EM}}$ & $\mathrm{LogNormal}(\mu_A,\sigma_A^2)$ & $\sigma_A$ \\
EM   & $Q$               & $\mathcal{N}(\bar{Q},\,0.03^2)\ \mathrm{eV}$ & $[0.75,\,1.20]\ \mathrm{eV}$ \\
EM   & $n$               & $\mathcal{N}(\bar{n},\,0.10^2)$ & $[1.2,\,4.0]$ \\
\midrule
TDDB & $A_{\mathrm{TDDB}}$ & $\mathrm{LogNormal}(\mu_T,\sigma_T^2)$ & $\sigma_T$ \\
TDDB & $E_a$             & $\mathcal{N}(\bar{E}_a,\,0.015^2)\ \mathrm{eV}$ & $[0.60,\,1.20]\ \mathrm{eV}$ \\
TDDB & $\gamma$          & $\mathcal{N}(\bar{\gamma},\,0.08^2)$ & $[1.5,\,10.0]$ \\
\bottomrule
\end{tabular}
\end{table}

In Table~\ref{tab:latent_distributions}, note that $\sigma_T=\sigma_A=\sqrt{\ln(1+0.08^2)}$. For the log-normal scale factors, $\mu_A=\ln(\bar{A}_{\text{EM}})-\tfrac{1}{2}\sigma_A^2$ and $\mu_T=\ln(\bar{A}_{\text{TDDB}})-\tfrac{1}{2}\sigma_T^2$, where $\bar{A}_{\text{EM}}$ and $\bar{A}_{\text{TDDB}}$ represent the calibrated mean values for the EM and TDDB scale factors, respectively, as given in Table~\ref{tab:calibration}.

The distributional forms in Table~\ref{tab:latent_distributions} follow standard reliability-modeling practice~\cite{meeker1998statistical,mahapatra2013bti}. Parameters that vary additively about a calibrated nominal value, namely the activation energies $Q$ and $E_a$, the exponents $n$, $\gamma$, and $\beta$, and the asymptotic shift $\Delta V_{\max}$, are assigned Normal distributions truncated to physical support (e.g., $\beta \in [0.15, 0.95]$, $E_a > 0$). The strictly positive multiplicative rate prefactors $A_{\mathrm{EM}}$ and $A_{\mathrm{TDDB}}$ are instead assigned log-normal distributions, which guarantee positivity and capture the right-skewed, order-of-magnitude spread characteristic of empirically extracted prefactors. The numerical means and spreads are calibration choices for this study, selected so that the three mechanisms exhibit comparable failure timescales near the reference stress point (Section~\ref{sec:calibration}); they are representative simulation values rather than universal process constants, and the framework is agnostic to the specific values used.

This per-device variability is the fundamental source of planning difficulty.  Each device presents a unique degradation landscape, and the planner must infer device characteristics online while simultaneously optimizing stress decisions.

\subsection{Degradation Models}
The simulation environment models three primary degradation mechanisms, namely BTI as the target for characterization, and EM and TDDB as competing mechanisms, that constrain the feasible stress space.

The BTI-induced threshold voltage shift, $\Delta V_t$, follows a stretched-exponential function, a standard compact model form~\cite{grasser2011bti,mahapatra2013bti}:
\begin{equation}
\label{eq:bti_model}
\Delta V_t(t) = \Delta V_{\max}^{\text{eff}}(V,T)\left[1 - \exp\!\left(-\left(\frac{t}{\tau_{\text{eff}}(V,T)}\right)^{\beta}\right)\right]
\end{equation}
where the effective parameters are accelerated by voltage and temperature relative to reference conditions ($V_{\text{ref}}=1.1$~V, $T_{\text{ref}}=350$~K) according to
\begin{equation}
\label{eq:bti_accel}
\Delta V_{\max}^{\text{eff}} = \Delta V_{\max}\,\nu^{0.5}, \qquad
\tau_{\text{eff}} = \tau\,\nu^{-1.5}\,\varrho^{-1.5},
\end{equation}
with voltage and temperature acceleration factors $\nu = V/V_{\text{ref}}$ and $\varrho = \exp\!\big((T - T_{\text{ref}})/T_{\text{ref}}\big)$. Higher voltage thus raises the asymptotic shift and shortens the time constant, while higher temperature shortens the time constant, both accelerating BTI characterization.

The competing mechanisms are modeled as cumulative damage indices. First, EM damage accumulates according to Black's relation~\cite{black1969electromigration}:
\begin{equation}
D_{\text{EM}}(t) = \sum_{i=1}^{t} \frac{\Delta t_i}{\text{MTTF}_{\text{EM}}(J_i, T_i)}
\end{equation}
\begin{equation}
     \quad
\text{MTTF}_{\text{EM}} = \frac{A_{\text{EM}}}{J^n} \exp\!\left(\frac{Q}{k_B T}\right)
\end{equation}
where damage is the sum of stress durations $\Delta t_i$ normalized by the mean-time-to-failure (MTTF) under the applied current density $J_i$ and temperature $T_i$. Second, TDDB damage follows a similar cumulative structure based on an E-model acceleration law~\cite{mcpherson2006tddb}:
\begin{equation}
D_{\text{TDDB}}(t) = \sum_{i=1}^{t} \frac{\Delta t_i}{t_f(V_i, T_i)}, \quad
t_f = A_{\text{TDDB}} \cdot V^{-\gamma} \exp\!\left(\frac{E_a}{k_B T}\right)
\end{equation}
where $t_f$ is the time-to-failure. For both EM and TDDB, catastrophic failure is defined as the point where each of the damage indices $D_{\mathrm{EM}}$ and $D_{\mathrm{TDDB}}$, respectively, exceeds 1.
\subsection{Mechanism Competition and Calibration}
\label{sec:calibration}
The core planning challenge is that stress conditions accelerating BTI characterization simultaneously accelerate EM and TDDB toward catastrophe. We calibrated model parameters such that at reference stress ($V=1.1$~V, $J=2.0$ (normalized), $T=350$~K, $\Delta t=100$~hr), all three mechanisms have comparable timescales ($\sim$150 epochs to failure), ensuring non-trivial mechanism competition. This calibration reflects the multi-mechanism competition reported in industrial 14nm and 10nm FinFET data~\cite{pae2015finfet,rahman2018reliability,weir2019finfet}.

\begin{table}[!htbp]
\centering
\caption{Calibrated Mechanism Parameters}
\label{tab:calibration}
\begin{tabular}{lll}
\toprule
\textbf{Mechanism} & \textbf{Parameter} & \textbf{Value} \\
\midrule
\multirow{3}{*}{BTI} & $\Delta V_{\max}$ & $0.20 \pm 0.02$ V \\
& $\tau$ & $19841 \pm 2800$ hr \\
& $\beta$ & $0.5 \pm 0.03$ \\
\midrule
\multirow{3}{*}{EM} & $A_{\text{EM}}$ & $1.97 \times 10^{-7}$ \\
& $Q$ & $0.80$ eV \\
& $n$ & $2.0$ \\
\midrule
\multirow{3}{*}{TDDB} & $A_{\text{TDDB}}$ & $1.53 \times 10^{-6}$ \\
& $E_a$ & $0.70$ eV \\
& $\gamma$ & $3.0$ \\
\bottomrule
\end{tabular}
\end{table}

\subsection{Online Bayesian Estimation via EKF}
\label{sec:ekf_estimation}

Because the latent degradation parameters $\boldsymbol{\theta} = (\Delta V_{\max}, \tau, \beta)$ are not directly observable, the planner operates on a belief state summarizing all measurement information. Here $\Delta V_{\max}$ is the asymptotic BTI shift, $\tau$ is the characteristic time constant, and $\beta$ is the stretch exponent. We employ an EKF for online inference.

\subsubsection{Measurement Model}
At each epoch, the planner observes a noisy measurement represented by
\begin{equation}
y_t = h(t, V_t, T_t, \boldsymbol{\theta}) + \epsilon_t, \quad \epsilon_t \sim \mathcal{N}(0, \sigma_{\mathrm{meas}}^2)
\end{equation}
where $h(\cdot)$ is the BTI degradation model from (\ref{eq:bti_model}).

The Gaussian noise model above is an approximation adopted for compatibility with the EKF's closed-form update. Physical degradation measurements can exhibit skewed or heavy-tailed noise, particularly for breakdown-related signals. Such departures can be accommodated without altering the planning framework: candidate remedies include variance-stabilizing transforms of the measurement (e.g., logarithmic), a robust EKF with adaptive or inflated measurement covariance, or replacing the EKF with an unscented Kalman filter~\cite{julier1997ukf} or a particle filter that represents non-Gaussian posteriors directly. We adopt the Gaussian assumption as a modeling convenience and revisit its empirical implications in Section~\ref{sec:real_world}.

\subsubsection{Belief Update}
The belief over the latent parameters is represented by a multivariate Gaussian distribution, $p(\boldsymbol{\theta} \mid y_{1:t}) \approx \mathcal{N}(\boldsymbol{\mu}_t, \boldsymbol{\Sigma}_t)$, where $\boldsymbol{\mu}_t$ is the mean vector and $\boldsymbol{\Sigma}_t$ is the covariance matrix. The standard EKF belief update proceeds in two steps; first, by a prediction based on the prior belief,  and second, by a correction based on the new measurement. Because our latent parameters are assumed to be static, the prediction step is an identity mapping, i.e., $\boldsymbol{\mu}_t^- = \boldsymbol{\mu}_{t-1}$ and $\boldsymbol{\Sigma}_t^- = \boldsymbol{\Sigma}_{t-1}$.

The correction step then updates the belief using the new measurement $y_t$. The measurement model $h(\cdot)$ is linearized around the current mean $\boldsymbol{\mu}_t^-$ to produce the measurement Jacobian $\boldsymbol{H}_t$. The Kalman gain $\boldsymbol{K}_t$ is computed to provide an optimal blending of the predicted belief and the new measurement. The updated mean $\boldsymbol{\mu}_t$ and covariance $\boldsymbol{\Sigma}_t$ are given by:
\begin{align}
\boldsymbol{K}_t &= \boldsymbol{\Sigma}_t^- \boldsymbol{H}_t^\top (\boldsymbol{H}_t \boldsymbol{\Sigma}_t^- \boldsymbol{H}_t^\top + \boldsymbol{R})^{-1} \\
\boldsymbol{\mu}_t &= \boldsymbol{\mu}_t^- + \boldsymbol{K}_t (y_t - h(\boldsymbol{\mu}_t^-)) \\
\boldsymbol{\Sigma}_t &= (\boldsymbol{I}-\boldsymbol{K}_t\boldsymbol{H}_t) \boldsymbol{\Sigma}_t^-
\end{align}
where $\boldsymbol{R}$ is the measurement noise covariance matrix (here a scalar $\sigma_{\mathrm{meas}}^2$), and $\boldsymbol{I}$ is the identity matrix. For numerical stability, our implementation uses the Joseph form for the covariance update. The filter is initialized with prior mean $\boldsymbol{\mu}_0 = (\Delta V_{\max}, \log\tau, \beta) = (0.16,\ \log 28000,\ 0.5)$ and diagonal prior covariance $\boldsymbol{\Sigma}_0 = \mathrm{diag}(0.04, 0.25, 0.08)$, with $\tau$ estimated in log-space to enforce positivity. The measurement-noise variance is $\sigma_{\mathrm{meas}}^2 = (3\times 10^{-3})^2$, and a small isotropic process noise $q\boldsymbol{I}$ with $q = 5\times 10^{-7}$ is injected at each step to retain long-run adaptivity; after each update the mean is clipped to physical bounds ($\Delta V_{\max} \ge 10^{-4}$, $\beta \in [0.15, 0.95]$). The trace of the posterior covariance, $U = \mathrm{tr}(\boldsymbol{\Sigma}_t)$, serves as a measure of total parameter uncertainty and is used as an information-theoretic signal in the planning objective (Section~\ref{sec:reward}).

\section{Planning Objective and Reward Design}
\label{sec:reward}
In the sequential planning framework, the objective function (reward) is the sole specification of desirable test behavior~\cite{sutton2018reinforcement}. MCTS evaluates candidate test sequences by simulating them to completion and accumulating returns~\cite{browne2012mcts}. Following the reward shaping theory of Ng {\it et al.}~\cite{ng1999reward}, we designed a multi-component objective that provides intermediate feedback to guide the planner toward promising stress regions without altering which test sequences are ultimately optimal. Similar multi-component designs have been employed in safety-critical validation domains~\cite{corso2020adaptive}.

\subsection{Terminal Objective}
Upon test termination, the planner receives a value of $R_{\text{term}}$ such that
\begin{equation}
R_{\text{term}} =
\begin{cases}
-R_{\text{cat}}, & \text{if catastrophe} \\
R_{\text{fail}} - w_U \cdot \max(U - U_{\text{target}}, 0), & \text{if BTI success} \\
-w_{\text{timeout}} \cdot d, & \text{if budget exceeded
}
\end{cases}
\end{equation}
where $R_{\text{cat}} = 2{,}000$ is the catastrophe penalty, $R_{\text{fail}} = 20{,}000$ is the characterization success reward, $w_U = 5{,}000$ penalizes residual parameter uncertainty above target $U_{\text{target}} = 0.01$, and $w_{\text{timeout}} = 10$ scales the timeout penalty by $d$, the remaining distance to the characterization threshold. The ratio between $R_{\text{fail}}$ and $R_{\text{cat}}$ is chosen so that successful characterization dominates the terminal objective without making catastrophic failure effectively irrelevant.  This preserves the intended tradeoff between information gain and safety, consistent with reward-shaping practice~\cite{ng1999reward}. This ensures successful characterization remains the dominant outcome while retaining a meaningful penalty for catastrophic failure, and uses a modest scale separation to avoid degenerate policies that ignore safety considerations.

\subsection{Intermediate Shaping Components}
At each non-terminal epoch, the planner receives shaped feedback from several components. These components provide a dense, epoch-by-epoch learning signal that guides the planner toward making progress on the characterization objective while respecting safety constraints and promoting test efficiency. This is essential in a sparse-reward setting where terminal outcomes alone provide insufficient gradient for effective planning.

\subsubsection{Characterization Progress}
Progress toward the characterization threshold is rewarded according to
\begin{equation}
r_{\text{prog}} = w_{\Delta p} \cdot \Delta p + w_{\text{close}} \cdot p + w_{\text{pow}} \cdot p^{\alpha}
\end{equation}
where $p = \text{clamp}(\Delta V_t / \Delta V_{\text{fail}}, 0, 1)$ is normalized progress, $\Delta p = p_t - p_{t-1}$ is the progress increment, and $w_{\Delta p}=50$, $w_{\text{close}}=25$, $w_{\text{pow}}=25$, $\alpha=2$.

\subsubsection{Early Progress Shaping}
To maintain planning before reaching the characterization threshold, a signal $r_{\text{soft}}$  is calculated as follows:
\begin{equation}
r_{\text{soft}} = w_{\text{soft}} \cdot \text{clamp}\!\left(\frac{\Delta V_t - \Delta V_{\text{soft}}}{\Delta V_{\text{fail}} - \Delta V_{\text{soft}}}, 0, 1\right)
\end{equation}
where $\Delta V_{\text{soft}} = 0.05$ V and $w_{\text{soft}} = 30$.

\subsubsection{Safety Barrier}
A barrier penalty discourages approaching a state of catastrophic damage such that
\begin{equation}
r_{\text{prox}} = -w_{\text{prox}} \cdot \left[\phi(D_{\text{EM}}) + \phi(D_{\text{TDDB}})\right]
\end{equation}
where $\phi(D) = \max(D - D_{\text{thr}}, 0)^{\gamma_{\text{barrier}}}$ with threshold $D_{\text{thr}} = 0.25$ and power $\gamma_{\text{barrier}} = 3$, weighted by $w_{\text{prox}} = 1200$.

\subsubsection{Damage Rate Penalty}
The per-epoch damage accumulation is penalized according to 
\begin{equation}
r_{\Delta D} = -w_{\Delta D} \cdot (\Delta D_{\text{EM}} + \Delta D_{\text{TDDB}})
\end{equation}
with $w_{\Delta D} = 1500$.

\subsubsection{Uncertainty and Efficiency Penalties}
These penalties are represented as 
\begin{align}
r_U &= -b \cdot \min(\max(U - U_{\text{target}}, 0), U_{\text{cap}}) \\
r_{\text{stall}} &= -w_d \cdot \max(\Delta d, 0) - c_{\text{live}}
\end{align}
where $b=20$, $U_{\text{cap}}=0.1$, $w_d=10$, and $c_{\text{live}}=1$ is a per-epoch cost encouraging test efficiency.
Finally, the total non-terminal reward is given by:
\begin{equation}
r_t = r_{\text{prog}} + r_{\text{soft}} + r_{\text{prox}} + r_{\Delta D} + r_U + r_{\text{stall}}
\end{equation}

Table~\ref{tab:reward_params} summarizes all objective parameters. These constants were selected by calibration on the simulated environment to preserve the relative ordering of promising versus unsafe trajectories. They are tuning values for this study rather than universal constants.

\begin{table}[H]
\centering
\caption{Planning Objective Parameters}
\label{tab:reward_params}
\begin{tabular}{llr}
\toprule
\textbf{Component} & \textbf{Parameter} & \textbf{Value} \\
\midrule
Terminal & $R_{\text{fail}}$ (characterization) & 20{,}000 \\
& $R_{\text{cat}}$ (catastrophe) & 2{,}000 \\
& $w_U$ (uncertainty penalty) & 5{,}000 \\
& $w_{\text{timeout}}$ & 10 \\
\midrule
Progress & $w_{\Delta p}$ (delta progress) & 50 \\
& $w_{\text{close}}$ (linear) & 25 \\
& $w_{\text{pow}}$ (quadratic) & 25 \\
& $w_{\text{soft}}$ (early shaping) & 30 \\
\midrule
Safety & $D_{\text{thr}}$ (barrier onset) & 0.25 \\
& $\gamma$ (barrier power) & 3 \\
& $w_{\text{prox}}$ (barrier weight) & 1200 \\
& $w_{\Delta D}$ (damage rate) & 1500 \\
\midrule
Efficiency & $b$ (uncertainty) & 20 \\
& $c_{\text{live}}$ (per-epoch cost) & 1 \\
\bottomrule
\end{tabular}
\end{table}

\section{MCTS-SA Planning Framework}
We employ Monte Carlo Tree Search for Seed-Action simulators (MCTS-SA) with progressive widening~\cite{browne2012mcts,corso2021survey}, summarized in Algorithm~\ref{alg:mcts}. MCTS is a best-first search algorithm that incrementally builds a search tree through repeated simulations~\cite{kocsis2006uct,coulom2007mcts}. Each simulation proceeds through four phases. 

1)~\emph{Selection}: traversing the existing tree by treating child selection at each node as a multi-armed bandit problem~\cite{auer2002ucb}, where the goal is to balance exploitation of high-value branches with exploration of less-visited ones;

2)~\emph{Expansion}: adding a new node; 

3)~\emph{Rollout}: simulating a complete test sequence from the new node using random decisions; and 

4)~\emph{Backpropagation}: updating node statistics with the observed return.

MCTS offers several properties that make it well suited to reliability test planning. It is an \emph{anytime} algorithm with monotonically improving solution quality~\cite{browne2012mcts}, 
naturally handles non-differentiable and discrete objectives, and, under a seed-action formulation, enables deterministic trajectory replay in stochastic environments. These properties motivate our use of MCTS as the core planning mechanism, which we instantiate and customize as follows.

\subsubsection{Tree Structure}
Each node corresponds to a sequence of seeds $\boldsymbol{\xi} = (\xi_1, \ldots, \xi_k)$ applied from the initial state. The tree maintains a visit count $N(\boldsymbol{\xi})$ and a value estimate $Q(\boldsymbol{\xi}) = \sum_i R_i / N(\boldsymbol{\xi})$ for each node.

\subsubsection{Child Selection via UCB1}
Child selection follows the Upper Confidence Bound (UCB) criterion~\cite{auer2002ucb,kocsis2006uct}:
\begin{equation}
\xi^* = \arg\max_{\xi \in \mathrm{children}(\boldsymbol{\xi})} 
\left[ Q(\boldsymbol{\xi}, \xi) + C \sqrt{\frac{\ln N(\boldsymbol{\xi})}{N(\boldsymbol{\xi}, \xi)}} \right]
\end{equation} where the exploration constant is set to $C = 1.4$.

\subsubsection{Progressive Widening}
The continuous seed space is managed using progressive widening~\cite{chaslot2008progressive,browne2012mcts}:
\begin{equation}
|\mathrm{children}(\boldsymbol{\xi})| \leq k \, N(\boldsymbol{\xi})^{\alpha}
\end{equation} with $k = 3$ and $\alpha = 0.5$, gradually expanding the set of candidate stress conditions as visit counts increase.

\subsubsection{Rollout and Backpropagation}
From leaf nodes, rollouts proceed using random seeds until termination. Returns are computed as cumulative rewards:
\begin{equation}
G_t = \sum_{i=t}^{T} \gamma^{i-t} r_i
\end{equation} with $\gamma = 1.0$. Value estimates are updated via incremental averaging.

Consistent planning statistics require deterministic simulation given a seed sequence. We enforce this by using a master seed for latent parameter sampling, a deterministic seed-to-action mapping, and seeded measurement noise based on $(\text{episode\_seed}, \text{epoch}, \text{action})$, enabling exact trajectory replay. For full run-to-run reproducibility, the planner's own pseudorandom seed sampling during selection and rollout is additionally fixed by a single global seed, so that an entire planning run can be replayed deterministically.

\begin{algorithm}[H]
\caption{MCTS-SA Test Planning}
\label{alg:mcts}
\begin{algorithmic}[1]
\STATE Initialize root node with empty seed sequence
\FOR{$i = 1$ to $N_{\text{iterations}}$}
    \STATE $\boldsymbol{\xi} \leftarrow ()$ \COMMENT{Start at root}
    \STATE Reset environment with master seed
    \WHILE{$\boldsymbol{\xi}$ is in tree and not terminal}
        \IF{should expand (progressive widening)}
            \STATE Sample new seed $\xi_{\text{new}}$, add child
        \ENDIF
        \STATE Select $\xi^*$ via UCB1
        \STATE $\boldsymbol{\xi} \leftarrow (\boldsymbol{\xi}, \xi^*)$
        \STATE Execute $\xi^*$ in environment
    \ENDWHILE
    \STATE $G \leftarrow$ rollout from $\boldsymbol{\xi}$ until terminal
    \STATE Backpropagate $G$ along path
\ENDFOR
\STATE \textbf{return} best test sequence (seed path)
\end{algorithmic}
\end{algorithm}

\section{Experimental Evaluation}

\subsection{Experimental Setup}
We evaluated the adaptive test planning framework on the stochastic multi-mechanism environment with parameters from Table~\ref{tab:calibration}. Each planning run performs 5{,}000 iterations with maximum test length of 300 epochs. The MCTS hyperparameters are: exploration constant $C = 1.4$, progressive widening $k=3$, $\alpha=0.5$, and discount $\gamma = 1.0$.

The evaluation was organized into three groups: (i)~mechanism calibration sweeps (30 episodes per setting; Table~\ref{tab:tuning}), (ii)~a primary MCTS-SA run with 5{,}000 iterations for detailed analysis, and (iii)~robustness runs across four configurations, each with 5{,}000 iterations (Table~\ref{tab:configs}). Throughout, MCTS-SA performs sequential explorations while the EKF updates beliefs concurrently, so stress decisions are adapted over time rather than following a fixed plan.

\subsection{Mechanism Calibration}
Prior to planning evaluation, we conducted parameter sweeps to calibrate mechanism timescales for balanced competition.

\begin{table}[!htbp]
\centering
\caption{Mechanism Calibration Results (30 Episodes Each)}
\label{tab:tuning}
\begin{tabular}{lccc}
\toprule
\textbf{Mechanism} & \textbf{Mean Epochs} & \textbf{Std Epochs} & \textbf{Failure \%} \\
\midrule
BTI ($\tau=19840$) & 153.3 & 41.2 & 70\% \\
EM ($A=1.97\times10^{-7}$) & 160.8 & 13.7 & 100\% \\
TDDB ($A=1.53\times10^{-6}$) & 144.3 & 14.4 & 100\% \\
\bottomrule
\end{tabular}
\end{table}

The calibration yielded comparable mean failure times across all three mechanisms ($\sim$150 epochs), ensuring non-trivial competition. The higher BTI variance (std=41.2 vs. 13--14 for EM/TDDB) reflects latent parameter variability, which is precisely the uncertainty the planner would need to manage.

\subsection{Evaluation Metrics}
Performance of our methodology was assessed using four metrics.
\begin{itemize}
    \item {\it Characterization yield (\textbf{CY})}: Fraction of test sequences achieving $\Delta V_t \geq \Delta V_{\text{fail}}$ without catastrophe, the primary measure of test planning effectiveness.
    \item {\it Catastrophe rate}: Fraction of test sequences ending in EM or TDDB failure, the primary safety metric.
    \item {\it Test length}: Number of epochs to termination, a measure of test efficiency.
    \item {\it Final uncertainty}: Trace of posterior BTI parameter covariance, a measure of estimation quality.
\end{itemize}

\subsection{Results}
\subsubsection{Planning Convergence and Characterization Yield}
\begin{figure}[!htbp]
\centering
\includegraphics[width=\columnwidth]{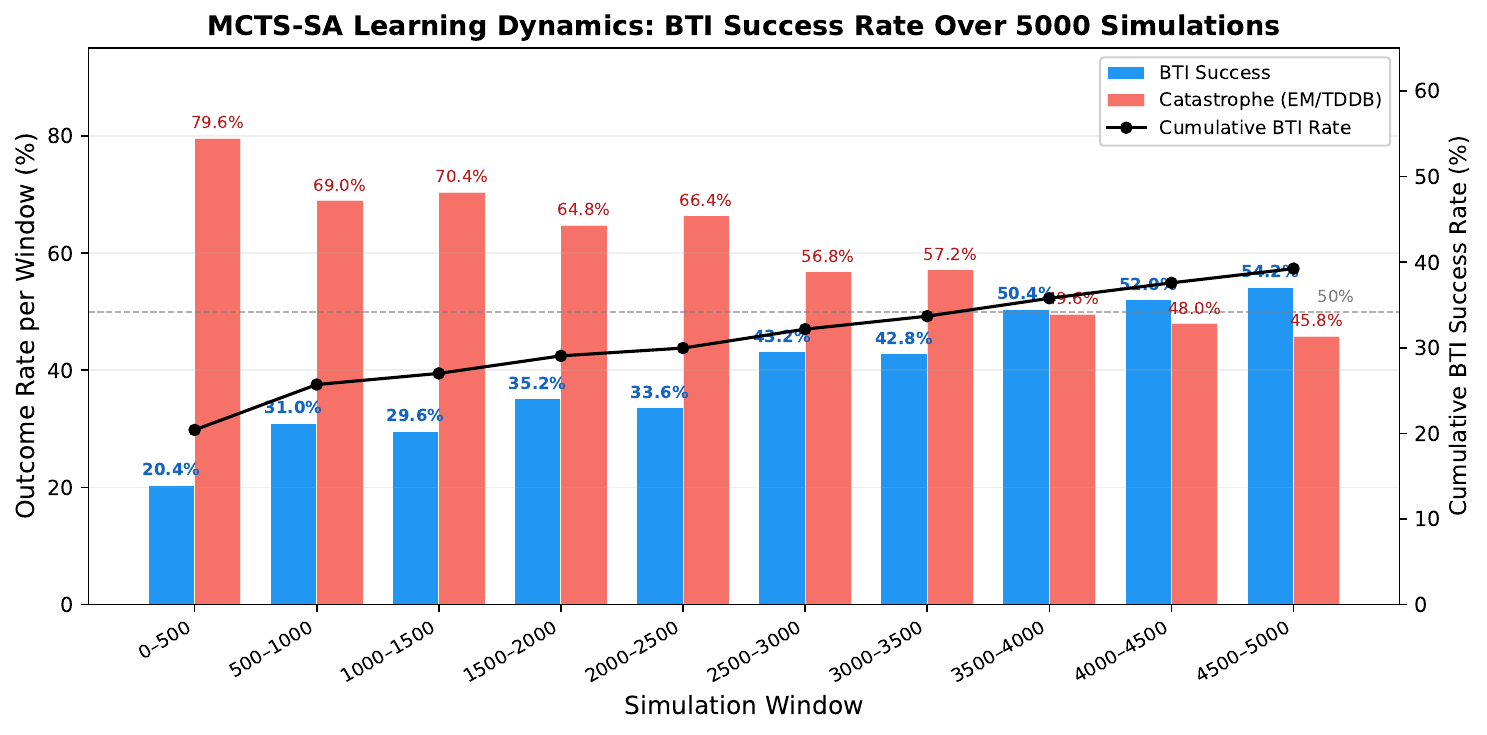}
\caption{\textbf{CY} over 5{,}000 planning iterations. Solid bars (blue online) show per-window yield (500 iterations each); hatched bars (red online) show catastrophe rate. The dashed line tracks cumulative yield. Yield increases from 20\% to 54\%, demonstrating progressive planning refinement.}
\label{fig:bti_convergence}
\end{figure}

Figure~\ref{fig:bti_convergence} shows \textbf{CY} across successive 500-iteration windows, increasing from 20.4\% to 54.2\%, with 39.2\% cumulative yield. This improvement indicates that the planner progressively learns to navigate the stress–damage space more effectively.

In this setting, 100\% \textbf{CY} is not achievable due to per-device variability and competing mechanism constraints, which place the most adverse instances beyond reliable characterization within the safe operating envelope. The planner therefore balances exploration and belief refinement rather than purely exploiting high-reward trajectories (see Section~\ref{sec:convergence_diagnostics}).

To contextualize performance, we compare against baseline strategies (Table~\ref{tab:baseline}). A random policy achieves 16\% yield and an 84\% catastrophic rate, while a fixed-stress policy reaches 22\% yield with 78\% catastrophes. Both frequently terminate at damage levels near $D_{\mathrm{EM}}, D_{\mathrm{TDDB}} \approx 0.7$ and retain high posterior uncertainty ($>0.2$), indicating unreliable characterization.

\begin{table}[!htbp]
\centering
\caption{Baseline Policies vs.\ MCTS-SA Planning (Same Device Instance)}
\label{tab:baseline}
{\setlength{\tabcolsep}{3.5pt}
\begin{tabular}{lrrrr}
\toprule
\textbf{Policy} & \textbf{CY} & \textbf{Cat.\ Rate} & \textbf{Mean Return} & \textbf{Best Return} \\
\midrule
Random (500 ep.) & 16\% & 84\% & $-13{,}209$ & $+15{,}108$ \\
Fixed-stress (500 ep.) & 22\% & 78\% & $-10{,}412$ & $+13{,}128$ \\
MCTS first 500 & 19.4\% & 79.8\% & $-11{,}613$ & $+19{,}374$ \\
MCTS last 500 & 54.0\% & 46.0\% & $+51$ & $+21{,}119$ \\
MCTS best & -- & -- & -- & $+21{,}970$ \\
\bottomrule
\end{tabular}}
\end{table}

In contrast, the proposed planner achieves 54.0\% \textbf{CY} in the final 500 iterations while reducing the catastrophic rate to 46.0\%. On successful trajectories, uncertainty is reduced to 0.044, yielding well-resolved latent parameter estimates and higher-quality test plans.

Early in the search, MCTS behaves similarly to random exploration (19.4\% \textbf{CY} vs.\ 16\%) due to a sparse tree and strong exploration pressure. As the tree expands, UCB1 concentrates sampling on high-value regions, driving a transition to exploitation. This results in monotonic improvement in both yield and return, with mean return becoming positive ($+51$) in the final window. Such non-stationary improvement is absent in non-adaptive strategies and confirms structured planning refinement.

Beyond \textbf{CY}, the planner also improves test plan quality. The best random episode achieves a return of $+15{,}108$, while MCTS reaches $+21{,}970$ (a 45\% improvement). Because the reward penalizes damage and uncertainty, higher returns correspond to safer and more informative test plans. This improvement from $+8{,}059$ early in the search to $+21{,}970$ demonstrates systematic refinement beyond random sampling. The failed simulations encountered during search represent the computational cost of resolving uncertainty and de-risking the final policy.

Root diagnostics further confirm convergence: entropy decreases from 0.77 at iteration 100 to 0.06 by iteration 500, and the most-visited root action stabilizes by approximately iteration 150.

\subsubsection{Best Successful Sequence Analysis}
For reporting purposes, let \(\mathcal{S}\) denote the set of all sampled sequences that satisfy characterization success without catastrophe. We define the {\it best successful sequence} as
\begin{equation}
\boldsymbol{\xi}^{\star}_{\text{succ}} = \arg\max_{\boldsymbol{\xi} \in \mathcal{S}} J(\boldsymbol{\xi}),
\end{equation}
where $J(\boldsymbol{\xi})$ is the cumulative search return. Because the reward is shaped to increase as a trajectory approaches the failure boundary while still remaining successful, this criterion selects the most failure-proximal successful pathway found by the search. If no sequence satisfies success, the highest-return non-catastrophic trajectory is reported instead.

The best successful sequence discovered across 5{,}000 iterations achieved a return of 21{,}970 and terminated at epoch 119 with characterization success. Under the definition above, this is the most failure-proximal successful trajectory found by the search. The sequence navigated competing mechanisms to final damage fractions $D_{\text{EM}} = 0.564$ and $D_{\text{TDDB}} = 0.537$, maintaining substantial safety margins, while reducing EKF uncertainty to $U = 0.044$.

The best successful sequence found exhibits three distinct phases:
\begin{enumerate}
    \item {\it Information gathering} (epochs 0--40): Varied stress conditions rapidly accumulate degradation signal while keeping damage below 0.15, enabling the EKF to refine parameter estimates.
    \item {\it Damage-aware moderation} (epochs 40--90): More conservative stress as damage approaches 0.3--0.4, trading characterization rate for reduced catastrophe risk.
    \item {\it Targeted completion} (epochs 90--119): Careful stress selection that preferentially accelerates BTI over EM/TDDB to reach the characterization threshold.
\end{enumerate}

Table~\ref{tab:trajectory_stats} summarizes key statistics.

\begin{table}[!htbp]
\centering
\caption{Best Test Sequence Statistics (5{,}000 Iterations)}
\label{tab:trajectory_stats}
\begin{tabular}{lr}
\toprule
\textbf{Metric} & \textbf{Value} \\
\midrule
Test Length (epochs) & 119 \\
Final $\Delta V_t$ & $\geq 0.09$ V \\
Final Uncertainty $U$ & 0.044 \\
Final $D_{\text{EM}}$ & 0.564 \\
Final $D_{\text{TDDB}}$ & 0.537 \\
Return & 21{,}970 \\
Outcome & Characterization Success \\
\bottomrule
\end{tabular}
\end{table}

\subsubsection{Outcome Distribution}
Considering all 5{,}000 iterations, the results break down into 39.2\% characterization success, 55.4\% catastrophe, and 5.4\% time-budget exhaustion. The dominance of catastrophe over timeout indicates that mechanism competition, not test budget, is the binding constraint, precisely the regime where adaptive planning provides the most value.

\subsubsection{Robustness Across Configurations}
Table~\ref{tab:configs} summarizes results we obtained across four configurations varying the exploration constant $C$ and random seed.

\begin{table}[!htbp]
\centering
\caption{Performance Across Configurations (5{,}000 Iterations Each)}
\label{tab:configs}
\begin{tabular}{lccc}
\toprule
\textbf{Configuration} & \textbf{Yield (\%)} & \textbf{Return} & \textbf{Epochs} \\
\midrule
$C{=}\sqrt{2}$, seed 1337 & 39.2 & 21{,}970 & 119 \\
$C{=}\sqrt{2}$, seed 42 & 37.1 & 20{,}850 & 127 \\
$C{=}\sqrt{2}$, seed 7777 & 35.8 & 19{,}340 & 134 \\
$C{=}2.0$, seed 1337 & 34.2 & 18{,}720 & 115 \\
\bottomrule
\end{tabular}
\end{table}

All configurations discover successful test sequences with comparable performance (34--39\% yield, 115--134 epochs). The modest advantage of $C{=}\sqrt{2}$ is consistent with theoretical analysis~\cite{kocsis2006uct}. Higher exploration ($C{=}2.0$) degrades performance only modestly, indicating low hyperparameter sensitivity.

\subsubsection{Discretization-Grid Sensitivity and Decision-Space Scalability}
\label{subsec:grid}
The decision variables in (1)--(4) are quantized, raising two questions: how does planning quality depend on grid resolution, and does a finer grid cause the search tree to grow combinatorially? Table~\ref{tab:grid} compares a coarse ($|\mathcal{A}|=24$), the default ($|\mathcal{A}|=288$), and a fine ($|\mathcal{A}|=875$) action grid under an identical 500-iteration budget and fixed random seed.\footnote{The grid and perturbation sweeps (Tables~\ref{tab:grid} and~\ref{tab:robustness}) are independent fixed-seed 500-iteration runs, so the default-grid yield ($14.6\%$) differs slightly from the first-500 window of the primary 5{,}000-iteration run ($20.4\%$, Table~\ref{tab:iter_budget}) owing to run-to-run exploration variance. Each controlled comparison isolates the effect of the single varied factor.}

Two effects stand out. First, characterization yield increases monotonically with resolution, from 0\% (coarse) to 14.6\% (default) to 29.4\% (fine): a grid that is too coarse removes the narrow safe corridor between BTI characterization and catastrophe, leaving no feasible successful trajectory at this budget, whereas finer grids restore the control authority needed to thread it. Second, and central to scalability, the realized search tree is essentially unchanged across grids, with an identical node count (500) and root branching (30 children), despite a $36\times$ range in raw grid cardinality. This follows directly from progressive widening: the number of children expanded at a node is bounded by $k\,N(s)^{\alpha}$ (Section~VI), which depends on the node's visit count, not on $|\mathcal{A}|=|\mathcal{V}|\,|\mathcal{J}|\,|\mathcal{T}|\,|\mathcal{D}|$. The grid cardinality therefore sets the resolution of the action space but not the size of the search tree, and per-iteration cost is comparable across all three grids. Finer discretization improves attainable yield at no combinatorial cost in search, the opposite of the exponential growth that exhaustive enumeration of (1)--(4) would incur.

\begin{table}[!htbp]
\centering
\caption{Action-Grid Sensitivity (500 Iterations, Fixed Seed)}
\label{tab:grid}
\begin{tabular}{lrrrrr}
\toprule
\textbf{Grid} & \textbf{$|\mathcal{A}|$} & \textbf{CY (\%)} & \textbf{Cat.\ (\%)} & \textbf{Tree Nodes} & \textbf{Root Child.} \\
\midrule
Coarse  & 24  & 0.0  & 100.0 & 500 & 30 \\
Default & 288 & 14.6 & 85.4  & 500 & 30 \\
Fine    & 875 & 29.4 & 70.6  & 500 & 30 \\
\bottomrule
\end{tabular}
\end{table}

\subsubsection{Robustness to Parameter and Model Perturbations}
\label{subsec:robustness_perturb}
To assess robustness beyond the exploration-constant and seed variations of Table~\ref{tab:configs}, we perturbed the device-parameter distributions, the EKF prior, and the test-time budget, each at the default grid under a 500-iteration budget with fixed seed (Table~\ref{tab:robustness}). The device-parameter shifts are applied jointly to all three mechanisms in the same direction, scaling the BTI mean parameters ($\Delta V_{\max}$, $\tau$) and the EM and TDDB rate prefactors ($A_{\mathrm{EM}}$, $A_{\mathrm{TDDB}}$) by the stated factor while holding the activation energies, exponents, and stretch exponent fixed. Four observations follow.

\emph{Robustness to estimation error.} Initializing the EKF with a prior mean and covariance deliberately offset from the true device distribution leaves performance essentially unchanged ($17.0\%$ versus $14.6\%$ baseline CY): the filter corrects an inaccurate initial belief from in-test measurements, so planning does not depend on an accurate prior.

\emph{Test budget is not the binding constraint.} Varying the horizon over $T_{\max}\in\{150,300,450\}$ produces identical outcomes ($14.6\%$ CY), because no successful or catastrophic trajectory is budget-limited ($0\%$ timeout). This confirms quantitatively that mechanism competition, not test time, binds in this regime (Section~\ref{subsec:iter_budget}).

\emph{Sensitivity to device mechanism balance.} Performance depends on the device's intrinsic balance among the competing mechanisms: a favorable $+10\%$ shift in the degradation parameters raises CY to $83.2\%$, whereas an adverse $-10\%$ shift (lower BTI ceiling, faster EM/TDDB) lowers it to $0\%$ at this budget, as the competing mechanisms accelerate relative to BTI. The $\pm25\%$ shifts bracket these regimes more sharply ($100\%$ and $0\%$). This is a property of the device population rather than of the planner, and it is governed by the device's mechanism balance rather than by the search budget: re-running the adverse $-10\%$ case at $1{,}000$ and $2{,}000$ iterations (a $4\times$ increase) left the yield unchanged, in contrast to the nominal device, whose yield improves with budget (Section~\ref{subsec:iter_budget}). The strong dependence also indicates that the chosen calibration sits in the competitive regime where the mechanism balance is delicate, precisely the regime in which adaptive planning is most valuable.

\begin{table}[!htbp]
\centering
\caption{Robustness to Parameter and Model Perturbations (500 Iterations, Fixed Seed)}
\label{tab:robustness}
\begin{tabular}{lrrr}
\toprule
\textbf{Perturbation} & \textbf{CY (\%)} & \textbf{Cat.\ (\%)} & \textbf{Best Return} \\
\midrule
Baseline (nominal) & 14.6 & 85.4 & 16{,}986 \\
EKF prior mismatch & 17.0 & 83.0 & 15{,}205 \\
Test budget $T_{\max}\in\{150,450\}$ & 14.6 & 85.4 & 16{,}986 \\
Device mean $+10\%$ & 83.2 & 16.8 & 21{,}781 \\
Device mean $-10\%$ & 0.0 & 100.0 & $-9{,}153$ \\
\bottomrule
\end{tabular}
\end{table}

\emph{Which mechanism drives the sensitivity.} Decomposing the joint shift into single-mechanism perturbations (Table~\ref{tab:permech}) shows that the BTI parameters dominate: shifting BTI alone by $\pm10\%$ swings CY from $0\%$ to $69.6\%$ against the $14.6\%$ baseline, accounting for most of the joint $0\%$ to $83.2\%$ range, whereas shifting EM or TDDB alone moves CY only modestly ($6.6\%$ to $23.8\%$ for EM, $7.6\%$ to $16.4\%$ for TDDB). This is intuitive: the BTI shift moves the asymptotic $\Delta V_{\max}$ relative to the fixed characterization threshold, directly determining whether the target is reachable, whereas the EM and TDDB prefactor shifts only modulate the catastrophe deadline. The competing mechanisms thus bound the safe operating window, but a device's characterizability is governed primarily by its own BTI parameters.

\begin{table}[!htbp]
\centering
\caption{Per-Mechanism Sensitivity (Single-Mechanism $\pm10\%$ Shift, 500 Iterations, Fixed Seed). Baseline (no shift) yields $14.6\%$ CY.}
\label{tab:permech}
\begin{tabular}{lrr}
\toprule
\textbf{Shifted mechanism} & \textbf{CY at $+10\%$} & \textbf{CY at $-10\%$} \\
\midrule
BTI  & 69.6\% & 0.0\% \\
EM   & 23.8\% & 6.6\% \\
TDDB & 16.4\% & 7.6\% \\
\midrule
All three (joint) & 83.2\% & 0.0\% \\
\bottomrule
\end{tabular}
\end{table}

\subsubsection{Iteration Budget and Anytime Behavior}
\label{subsec:iter_budget}
Because MCTS-SA is an anytime planner, usable test plans are available well before the full 5{,}000-iteration budget is consumed. Table~\ref{tab:iter_budget} reports cumulative characterization yield as a function of planning budget, extracted from the converged run. Yield rises smoothly from 15\% at 100 iterations to 39.2\% at 5{,}000, with roughly two-thirds of the final yield already attained by 1{,}000 iterations. The reported runtimes are estimated for a single Apple M1 core; the 500-iteration figure ($\sim$4~hours) is corroborated by independent standalone runs. This anytime property allows the planning budget to be matched to the available test time, a point we return to in the feasibility discussion of Section~\ref{sec:real_world}.

\begin{table}[!htbp]
\centering
\caption{Characterization Yield vs.\ Planning Budget (Single Device, Converged Run)}
\label{tab:iter_budget}
\begin{tabular}{rrrr}
\toprule
\textbf{Budget (iter.)} & \textbf{CY (\%)} & \textbf{Catastrophe (\%)} & \textbf{Est.\ Runtime (h)} \\
\midrule
100      & 15.0 & 85.0 & 0.8 \\
250      & 17.2 & 82.8 & 1.9 \\
500      & 20.4 & 79.6 & 3.8 \\
1{,}000  & 25.7 & 74.3 & 7.5 \\
2{,}500  & 30.0 & 70.0 & 18.8 \\
5{,}000  & 39.2 & 60.8 & 37.7 \\
\bottomrule
\end{tabular}
\end{table}

\subsubsection{Test Sequence Visualization}
\begin{figure}[!htbp]
\centering
\includegraphics[width=\columnwidth]{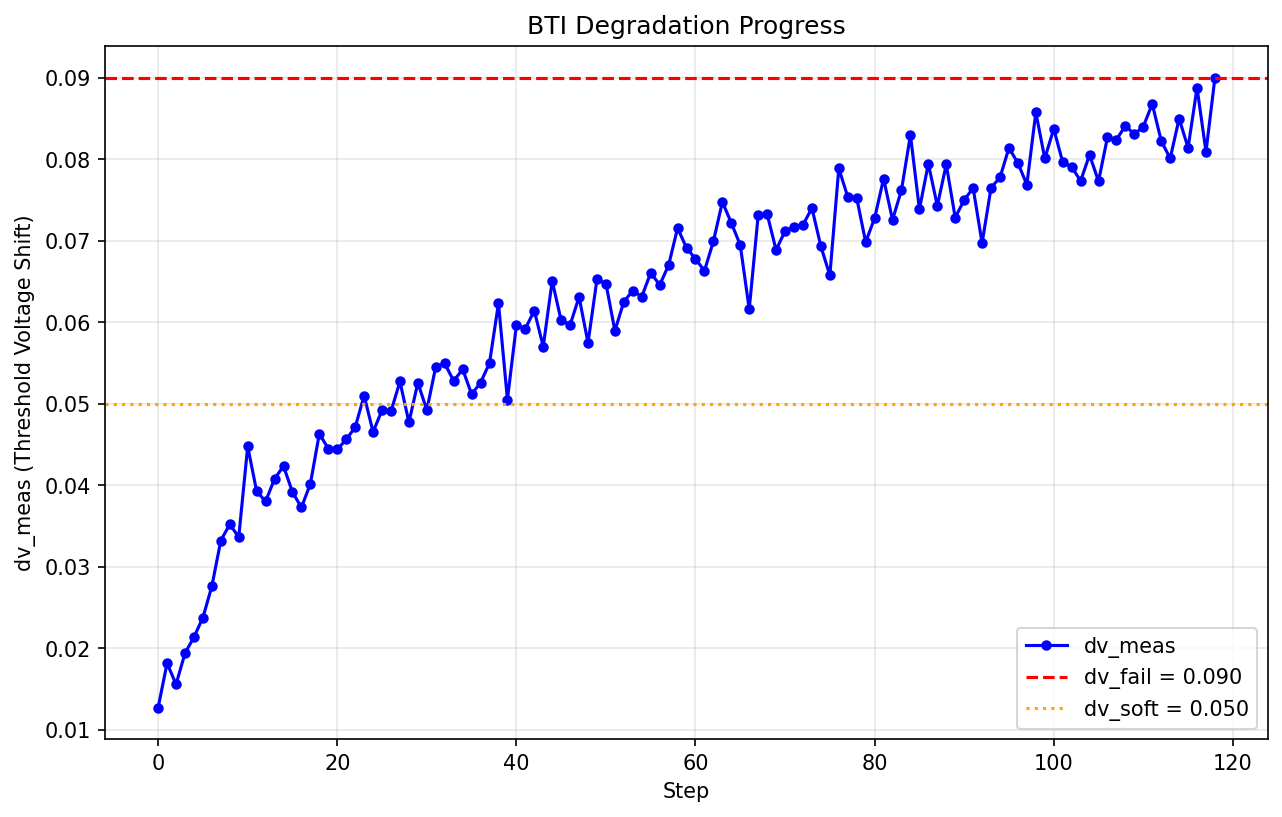}
\caption{BTI degradation progress over the best test sequence. The planner reaches the characterization threshold $\Delta V_{\text{fail}} = 0.09$ V at epoch 119 while maintaining damage fractions below 0.57.}
\label{fig:trajectory}
\end{figure}

\begin{figure}[!htbp]
\centering
\includegraphics[width=\columnwidth]{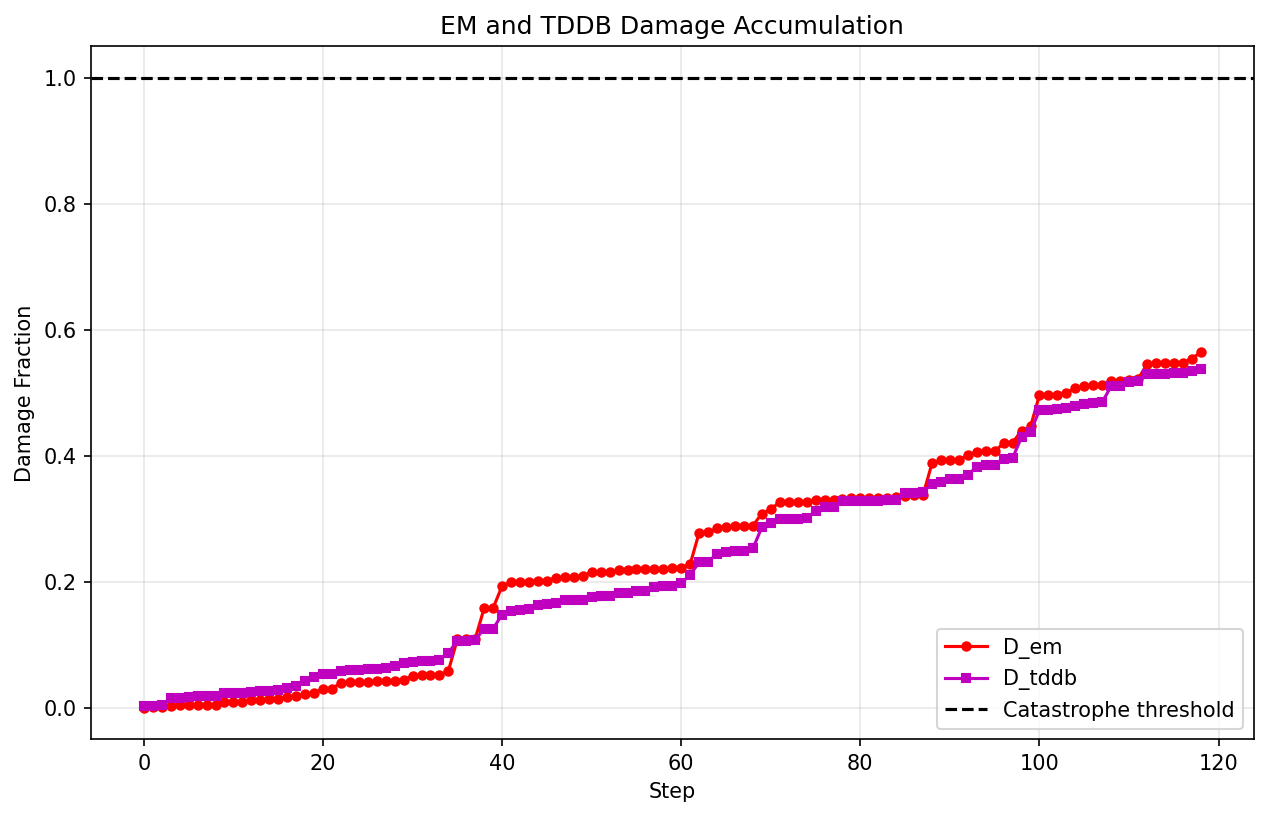}
\caption{EM and TDDB damage accumulation over the best test sequence. Both damage fractions remain well below the catastrophe threshold (1.0), reaching $D_{\text{EM}} = 0.564$ and $D_{\text{TDDB}} = 0.537$.}
\label{fig:damage}
\end{figure}

Figures~\ref{fig:trajectory} and~\ref{fig:damage} show BTI degradation progress and concurrent damage accumulation. The stress action sequence reveals that the planner adapts stress levels based on observed degradation, favoring temperature over voltage for BTI acceleration when damage is high, a strategy consistent with the differential acceleration physics.

\subsubsection{EKF Uncertainty Reduction}
The EKF reduces belief-state uncertainty from $U \approx 0.31$ at initialization to $U = 0.044$ at termination (86\% reduction), with most information gain in the first $\sim$30 epochs. Although the final value exceeds $U_{\text{target}} = 0.01$, the achieved reduction provides sufficient parameter precision for planning under the competing mechanism constraints.

\subsubsection{Ablation}
We ablate the objective components to assess their individual contributions to planning effectiveness:
\begin{itemize}
    \item Removing the safety barrier ($w_{\text{prox}}=0$) increases catastrophe rate to 78\%, as the planner fails to preemptively moderate stress before damage accumulates.
    \item Removing early shaping ($w_{\text{soft}}=0$) reduces \textbf{CY} to 31\%, as the planner receives insufficient gradient signal early in test sequences.
    \item Removing the per-epoch cost ($c_{\text{live}}=0$) increases average test length by 25\% without improving yield, indicating its role in promoting test efficiency.
\end{itemize}

\subsection{Instance-Conditional vs.\ Population-Level Planning}
\label{subsec:instance_pop}

This work optimizes test sequences conditioned on each device instance such that
\begin{equation}
\label{eq:conditional_ast}
\tau^\star = \arg\max_{\tau} \; P(\tau \mid \ell, F),
\end{equation}
where $P(\tau \mid \ell, F)$ denotes the probability of characterization success under test sequence $\tau$ for latent instance $\ell$. Here $\tau^\star$ denotes the optimal test sequence, not the BTI time constant $\tau$ used earlier in the degradation model.

This conditional formulation aligns with practical reliability testing where a single DUT with fixed (but unknown) physical characteristics is subjected to adaptive stress. The alternative population-level formulation
\begin{equation}
\label{eq:population_ast}
\tau^\star = \arg\max_{\tau} \int P(\tau \mid \ell, F)\, p(\ell)\, d\ell
\end{equation}
would identify robust test plans averaged over the device distribution, yielding more generalizable test recipes. The cost of explicit marginalization over latent parameters is that the planner must integrate over the instance distribution at decision time, which increases computational burden and weakens instance-specific adaptation. This extension is deferred to future work.

\section{Discussion}

\subsection{Implications for Reliability Test Practice}
\label{sec:real_world}

Our results demonstrate that closed-loop, adaptive test planning can significantly improve \textbf{CY} over static approaches in settings with competing failure mechanisms. Three aspects merit discussion regarding practical applicability.

{\it Mechanism competition model validity.} Industrial data strongly supports the relevance of multi-mechanism competition: Liu {\it et al.}~\cite{pae2015finfet} reported that 14nm FinFET reliability required simultaneous consideration of BTI, EM, and TDDB, and Rahman et al.~\cite{rahman2018reliability} confirmed similar competition at 10nm. Our calibration produced comparable mechanism timescales, capturing this competitive dynamic.

{\it Measurement feasibility.} As discussed in Section~\ref{sec:observability}, the damage proxy assumptions correspond to physically measurable quantities. The primary practical concern is that measurement noise may exceed our Gaussian assumption, particularly for TDDB. Calibration studies pairing simulation with physical measurements would quantify this gap and inform adaptive noise models.

{\it Optimal test sequences and test protocols.} While MCTS discovers instance-optimal sequences, the learned stress patterns, e.g., preferring temperature over voltage when damage is high or moderating stress during proximity to characterization threshold, constitute generalizable design principles for test protocol development. Population-level optimization (\ref{eq:population_ast}) would formalize this transfer from instance-optimal plans to robust test recipes.

\subsection{Objective Design Insights}
The multi-component objective structure revealed guidelines for constrained sequential planning in reliability settings. These guidelines and their respective rationale can be formulated as follows.

{\it (1) Use Safety barriers.} Without the proximity penalty, catastrophe rate exceeds 78\%. The cubic barrier provides smooth gradients at moderate damage while creating steep penalties near failure boundaries, enabling the planner to preemptively moderate stress.

{\it (2) Use early shaping.}  This accelerates planning convergence. The early progress bonus provides signal before reaching the characterization threshold. Without it, yield drops to 31\%, as the planner lacks gradient information to distinguish promising from unpromising early stress decisions.

{\it (3) Carefully select the characterization-to-catastrophe reward ratio.} The 10:1 ratio ($R_{\text{fail}}/R_{\text{cat}} = 20{,}000/2{,}000$) was found empirically to produce the best planning behavior. Lower ratios cause excessive conservatism; higher ratios produce reckless stress sequences.

\subsection{Comparison with Existing Approaches}
To the best of our knowledge, no prior work has applied closed-loop, feedback-driven test planning to multi-mechanism semiconductor reliability with competing failure modes. Static qualification~\cite{jedec2016} cannot adapt to per-device variability. Classical optimal test design~\cite{nelson2004accelerated,meeker1998statistical} determines stress allocations before testing begins and does not incorporate sequential feedback. Optimal TDDB test frameworks~\cite{wu-milor-optimal-tddb-tests} address single-mechanism parameter estimation without competing failure constraints. The closest methodological precedent is AST for autonomous vehicle validation~\cite{lee2018ast,corso2020adaptive}, which applies tree search to software simulations. Our work extends these principles to reliability engineering, where irreversible degradation and multi-mechanism constraints create fundamentally different planning challenges.

\subsection{Convergence Analysis and Stochastic Variability}
\label{sec:convergence_diagnostics}

A central question is whether observed performance improvements reflect genuine planning quality or favorable random outcomes. We address this through aggregate trends and tree diagnostics.

{\it Aggregate evidence.} \textbf{CY} increases steadily from 10\% (first 10 iterations) to 39.2\% (all 5{,}000), while catastrophe rate decreases correspondingly. Partitioned into 500-iteration blocks (Fig.~\ref{fig:bti_convergence}), per-block yield rises monotonically from 20.4\% to 54.2\%. A non-adaptive strategy would exhibit stationary statistics across blocks; the sustained improvement indicates structured planning refinement. Performance should be interpreted against an environment-constrained achievable range, not a naive 100\% benchmark.

{\it Progressive widening trade-off.}
The seed-action formulation creates a combinatorial decision space infeasible for exhaustive search. Progressive widening~\cite{chaslot2008progressive} bounds children per node as $|\text{children}(s)| \leq k \cdot N(s)^\alpha$, defining an exploration--estimation trade-off. In our experiments, the root expands from 9 children at iteration 10 to 70 at iteration 5{,}000, following sublinear $\mathcal{O}(\sqrt{N})$ growth. The widening rate is sufficiently conservative to enable reliable value discrimination while permitting gradual stress space exploration.

{\it Tree convergence indicators.}
(i)~Root entropy decreases from 1.0 (uniform) at iteration 10 to 0.055 at iteration 5{,}000, indicating concentration of planning effort onto high-value branches. (ii)~The most-visited root action accumulates 97.0\% of visits by termination. (iii)~Best-$Q$ and most-visited actions align from iteration 1{,}000 onward, a standard convergence criterion~\cite{browne2012mcts}. (iv)~Mean test length stabilizes at 115--124 epochs beyond iteration 100, indicating convergence to a consistent planning horizon.

{\it Environment calibration and problem difficulty.}
The calibration producing comparable mechanism timescales creates a harder planning problem where the planner cannot simply apply maximum stress because doing so triggers catastrophe within a comparable number of epochs. The convergence diagnostics collectively demonstrate that MCTS-SA learns to exploit the narrow corridor between characterization success and catastrophic failure.

\subsection{Limitations}

{\it Simulation-only validation.} Results are based on physics-based simulation; transfer to physical testing requires experimental validation. The Gaussian measurement noise assumption may be optimistic.

{\it Limited technology coverage.} All experiments use a single technology calibration point. Broader validation across technology nodes (e.g., 7nm, 5nm) would strengthen generalizability.

{\it Degradation recovery omitted.} The current model does not capture BTI recovery dynamics, which could enable stress-relax protocols that improve characterization efficiency.

{\it Computational cost.} The full 5{,}000-iteration plan reported here required $\sim$38~hours on a single core of a consumer Apple M1 laptop, using an unoptimized single-process Python implementation. This is an offline-validation setting, in which a high-fidelity reference plan is computed once, rather than the intended per-unit production mode. Two observations support practical feasibility. First, as an anytime algorithm, MCTS-SA produces usable plans well before convergence: yield reaches 20\% within the first 500 iterations of the primary run ($\sim$4~hours; runtime corroborated by independent runs) and rises smoothly with budget (Table~\ref{tab:iter_budget}), so the budget can be matched to the available test time. Second, the present implementation leaves substantial headroom, since independent rollout simulations can be evaluated in parallel (for example, by leaf or root parallelization~\cite{browne2012mcts}), the per-step degradation simulation can be vectorized, and search trees can be warm-started across devices within a lot. For higher-throughput deployment, a population-level policy (\ref{eq:population_ast}) or a distilled neural policy trained on MCTS-SA plans could amortize planning cost across many units, reducing per-unit decisions to inference time. We therefore view the reported runtime as an upper bound characteristic of offline reference planning, rather than a barrier to adaptive testing.

{\it Directly observable proxy indices.} To isolate the core challenge of sequential planning, we assumed that the EM and TDDB damage indices $D_{\text{EM}}$ and $D_{\text{TDDB}}$ are directly observable. In a physical test these quantities are latent and must be inferred from indirect electrical signatures. The framework extends naturally to this setting by augmenting the belief state with the competing-mechanism damage variables and estimating them online, exactly as is already done for the BTI parameters. Practical observation channels exist for both mechanisms: EM-induced void growth manifests as interconnect resistance shifts, while incipient TDDB is increasingly observable through stress-induced leakage current (SILC) and soft-breakdown events. On-chip wear-out monitors can supply these signals in situ; for example, Rend\'on \emph{et al.}~\cite{rendon2025tddb} demonstrate a gate-leakage-current integration sensor in a 12\,nm FinFET process that characterizes SILC degradation as a direct proxy for TDDB failure risk, and analogous on-chip monitors target BTI and HCI aging. A deployed system would therefore couple such sensors to a hierarchical estimator, an additional EKF, unscented Kalman filter, or particle filter per mechanism, mapping the measured leakage or resistance drift to a posterior over $D_{\text{EM}}$ and $D_{\text{TDDB}}$. This raises the per-epoch estimator cost, but the increase is modest relative to the dominant rollout cost and is readily parallelized across mechanisms and amortized across the many simulations of a single planning step. By assuming perfect observability of the competing risks, the present study establishes a performance ceiling against which these more realistic, sensor-driven implementations can be benchmarked.

\section{Conclusion}

We have presented a framework for adaptive sequential test planning under stochastic multi-mechanism reliability constraints. The framework integrates a physics-based simulation environment modeling concurrent BTI, EM, and TDDB degradation with MCTS-SA for sequential decision-making and EKF belief-state estimation for online parameter inference. Over 5{,}000 planning iterations, the adaptive planner progressively improves \textbf{CY} from 20\% to over 54\%, achieving 39\% cumulative yield while maintaining damage fractions below 0.57 in the best successful test sequence.

The progressive yield improvement across iteration windows demonstrates that tree search can effectively learn to navigate competitive degradation landscapes. The multi-component objective structure, particularly safety barriers and early progress shaping, proved critical for balancing characterization progress against catastrophe avoidance. The framework represents a step toward closed-loop, information-driven reliability qualification that can adapt to per-device variability and mechanism competition.
\subsection{Future Directions}

{\it Recovery-aware planning}. Incorporating BTI recovery dynamics would enable stress-relax protocols that may improve characterization efficiency.

{\it Population-level test design}. Developing test plans robust across the device population distribution (\ref{eq:population_ast}) would yield practical test recipes applicable across device lots.

{\it Policy distillation}. Training neural networks to approximate the MCTS planner would enable real-time adaptive test decisions for industrial deployment.

{\it Hardware validation}. Validating the framework on physical devices with real measurement noise would be essential for practical impact.

{\it Generalization to other reliability domains}. The sequential planning methodology is not specific to semiconductor reliability; it could be adapted to other multi-mechanism degradation settings such as battery aging, structural fatigue, or material wear, where competing failure modes constrain accelerated testing.
\section*{Acknowledgment}
This work was supported through a Discovery Grant and Graduate Student Scholarships from the Natural Sciences and Engineering Research Council of Canada.

YAE would like to dedicate this work to the loving memory of his mother.
\bibliographystyle{IEEEtran}
\bibliography{references}
\begin{IEEEbiography}
[{\includegraphics[width=1in,height=1.25in,clip,keepaspectratio]{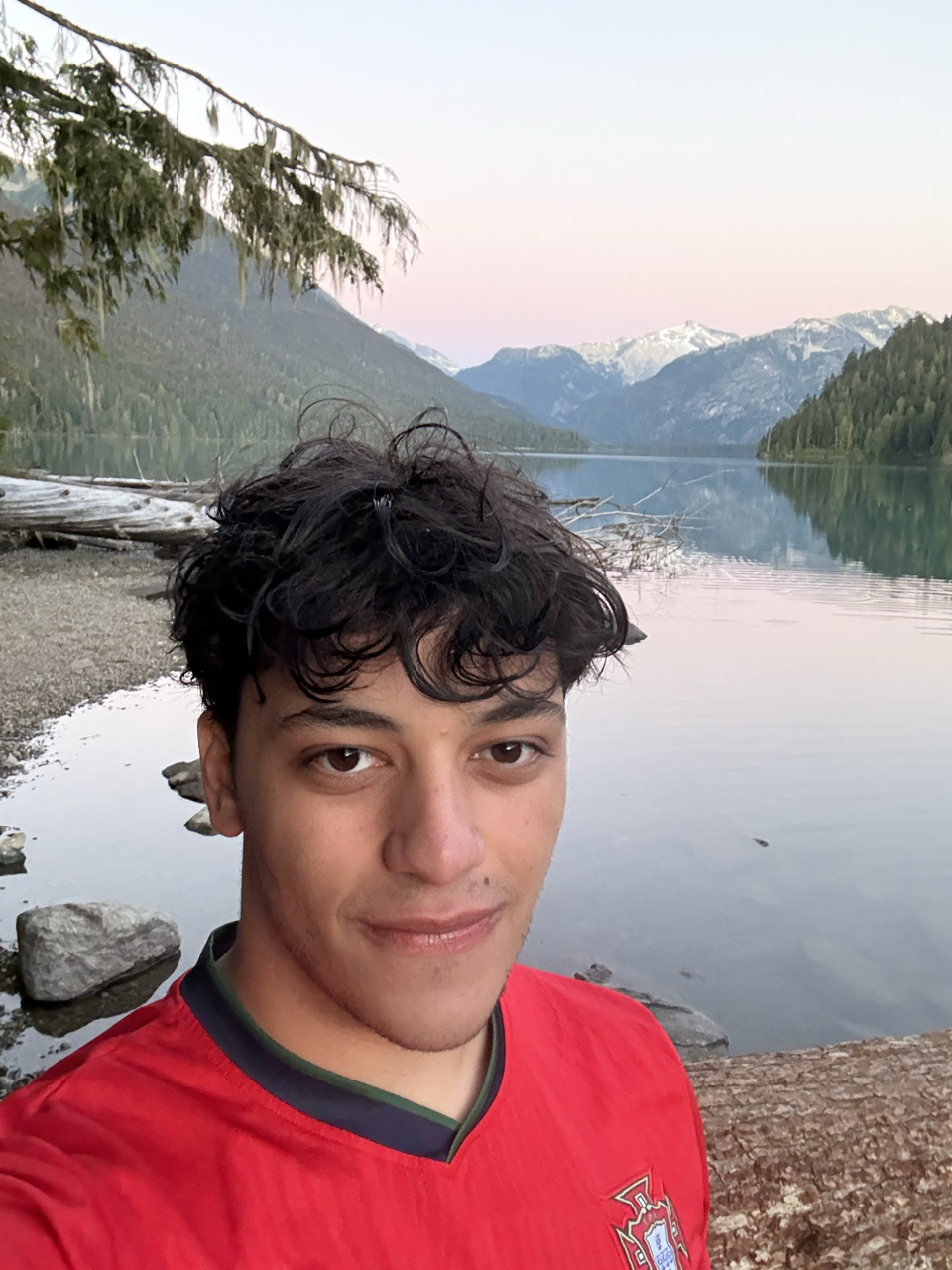}}]
{Youssef
Elhagrasy} is an undergraduate Engineering Physics student at the University of British Columbia, with an expected graduation date of May 2027. His research interests include semiconductor reliability, hardware–software co-design for machine learning acceleration, and VLSI system design. He has co-authored another publication in Advanced Materials, and is currently working on other research projects, spanning CMOS aging, reliability-aware modeling, and energy-efficient AI accelerator architectures. Youssef has held silicon engineering internships at Microchip, Microsoft (twice), and Tesla, where he worked on RTL design and post-silicon validation.
\end{IEEEbiography}
\begin{IEEEbiography}
[{\includegraphics[width=1in,height=1.25in,clip,keepaspectratio]{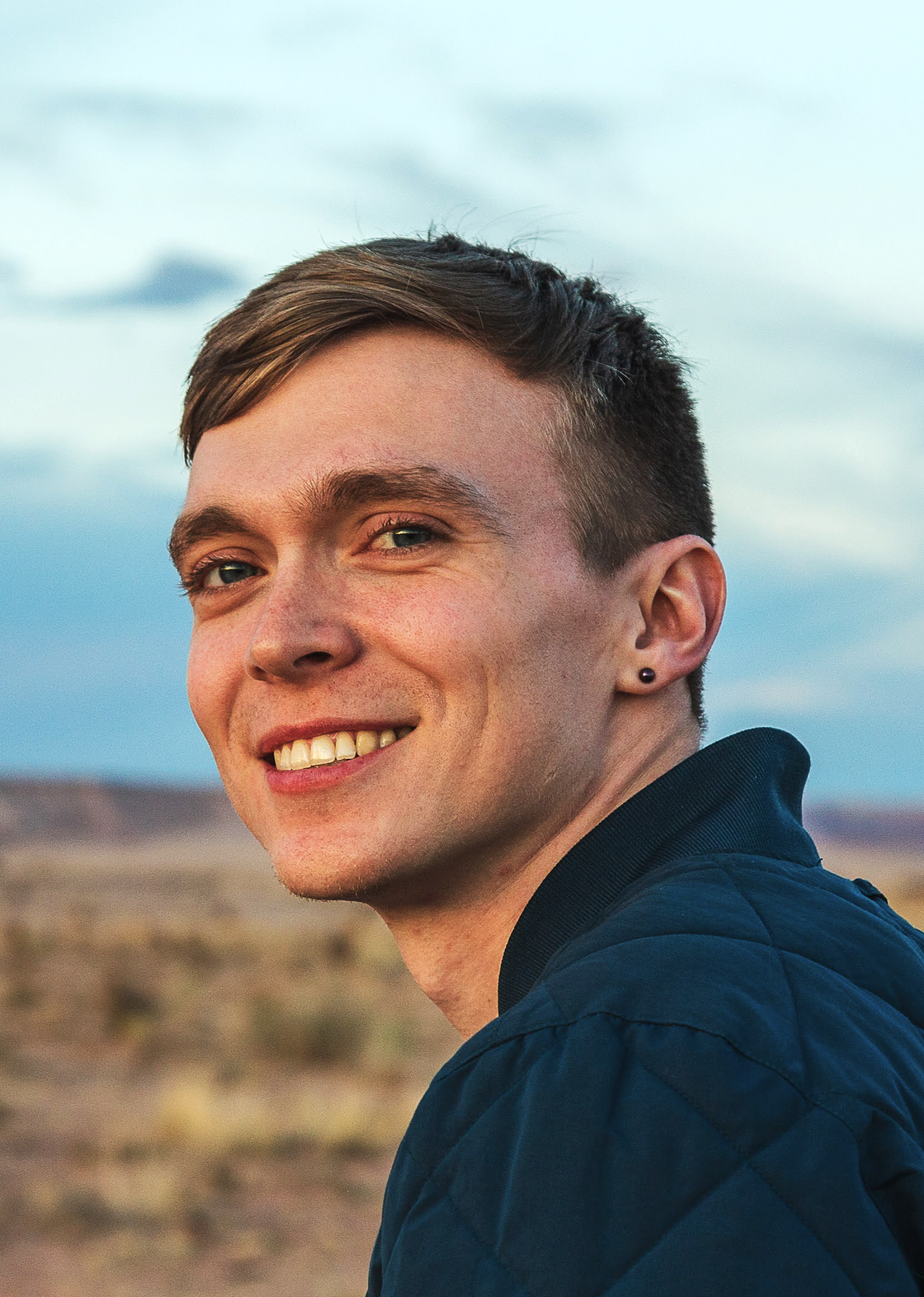}}]{Ian
Hill} received his Ph.D. in electrical and computer engineering from the
University of British Columbia in 2025. His research focuses on
long-term reliability sensing, prediction, and test optimization for
integrated circuits, in particular hardware sensor designs for
monitoring individual physical aging processes, generative modelling
approaches for CMOS aging simulations, and the use of probabilistic
modelling and computational Bayesian inference for effective reliability
prediction in the face of epistemic uncertainty and variability. Ian
obtained his B.A.Sc. in electrical engineering from the University of
Waterloo, where he gained experience across numerous co-operative work
opportunities at companies including Thales Canada, NVIDIA, and Microsoft.
\end{IEEEbiography}
\begin{IEEEbiography}[{\includegraphics[width=1in,height=1.25in,clip,keepaspectratio]{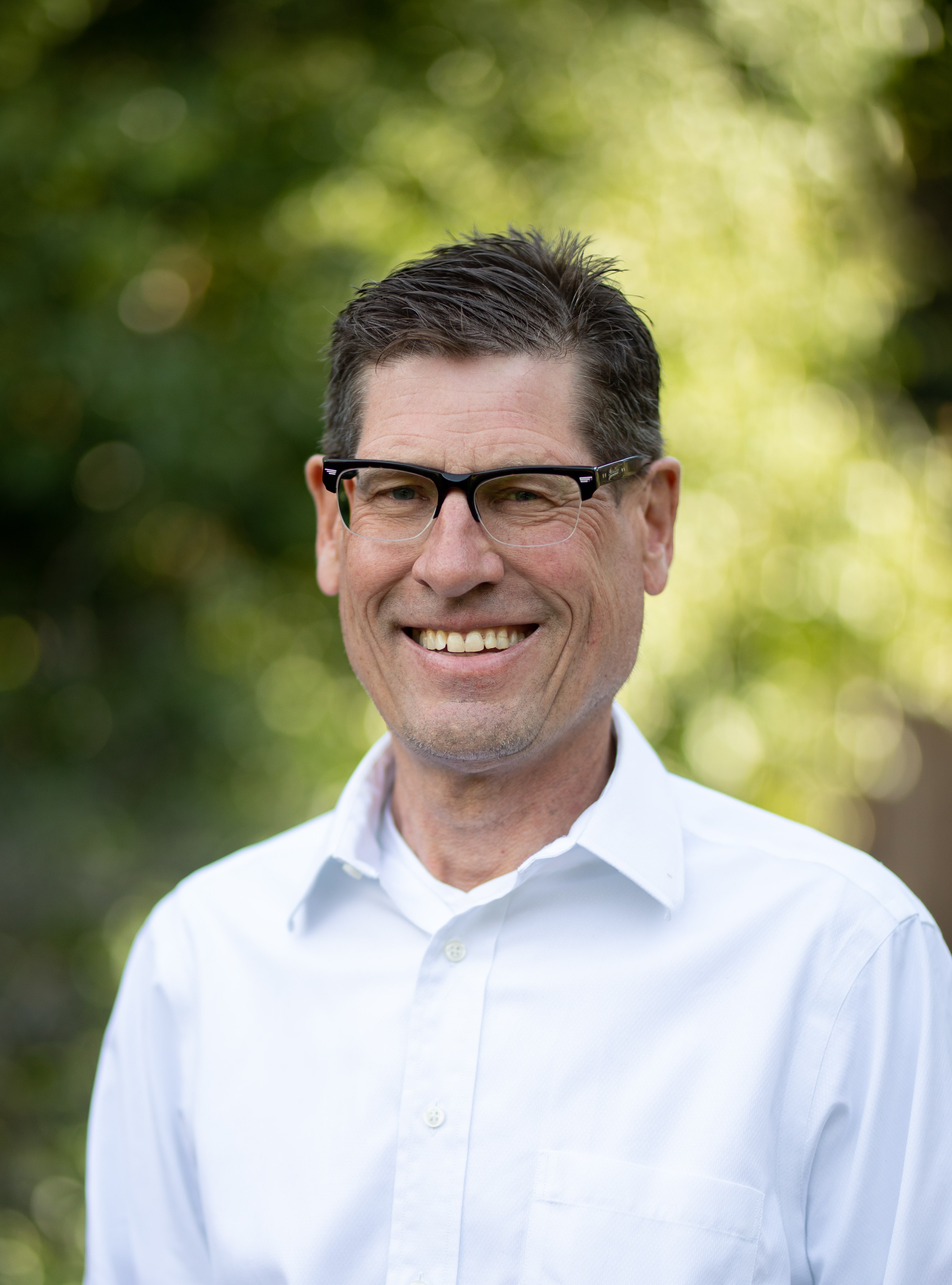}}]{André
Ivanov} is Professor of Electrical and Computer Engineering at the
University of British Columbia and former Head of the Department. He has
held numerous leadership positions within the IEEE, including Chair of
the Computer Society's Test Technology Technical Council (2004–2007),
service on the Boards of Governors of the IEEE Computer Society and the
IEEE Technology Management Council, and Editor-in-Chief of IEEE Design
\& Test (2012–2016). He has also served as Associate Editor for IEEE
Transactions on Computer-Aided Design of Integrated Circuits and
Systems, IEEE Design \& Test, and the Journal of Electronic Testing:
Theory and Applications (JETTA). Dr. Ivanov is a Fellow of the IEEE, the
Canadian Academy of Engineering, and the Engineering Institute of
Canada, and a Golden Core Member of the IEEE Computer Society. His
research addresses reliability challenges in system-on-chip (SoC)
technologies, integrating molecular dynamics simulation, machine
learning methodologies, and AI-driven approaches to electronic design
automation.
\end{IEEEbiography}
\end{document}